\documentclass[letterpaper, 10 pt, conference]{ieeeconf}  
\usepackage{color}
\usepackage{graphicx} 
\usepackage{amssymb}
\usepackage{amsmath}
\usepackage{algorithm} 
\usepackage{epsfig}
\usepackage{epstopdf}
\usepackage{multirow}
\usepackage{algorithmic}
\usepackage[algo2e]{algorithm2e}
\usepackage{subfigure}
\usepackage{bm}
\usepackage{hyperref}
\usepackage{cite}

\newtheorem{definition}{Definition}[section]

\IEEEoverridecommandlockouts                              

\title{\LARGE \bf
Measuring Similarity of Interactive Driving Behaviors Using Matrix Profile}

\author{Qin Lin$^{1}$, Wenshuo Wang$^{2}$,~\IEEEmembership{Member,~IEEE}, Yihuan Zhang$^{3}$, and John M. Dolan$^{1}, ~\IEEEmembership{Senior Member,~IEEE}$
\thanks{$^{1}$Qin Lin and John Dolan are with the Robotics Institute, Carnegie Mellon University, Pittsburgh, PA 15213, USA {\tt\small qinlin,jdolan@andrew.cmu.edu}}%
\thanks{$^{2}$Wenshuo Wang is with the California Partners for Advanced Transportation Technology (PATH), University of California, Berkeley, CA 94570. He was with Carnegie Mellon University, Pittsburgh, PA 15213,
USA. {\tt\small wwsbit@gmail.com}}%
\thanks{$^{3}$Yihuan Zhang is with Tsinghua Automotive Research Institute (Suzhou), China. {\tt\small 13yhzhang@tongji.edu.cn}}%
}
\begin{document}

\maketitle
\thispagestyle{empty}
\pagestyle{empty}

\begin{abstract}
Understanding multi-vehicle interactive behaviors with temporal sequential observations is crucial for autonomous vehicles to make appropriate decisions in an uncertain traffic environment. On-demand similarity measures are significant for autonomous vehicles to deal with massive interactive driving behaviors by clustering and classifying diverse scenarios. This paper proposes a general approach for measuring spatiotemporal similarity of interactive behaviors using a multivariate matrix profile technique. The key attractive features of the approach are its reduced space and time complexity, real-time online computing for streaming traffic data, and possible capability of leveraging hardware for parallel computation. The proposed approach is validated through automatically discovering similar interactive driving behaviors at intersections from sequential data.
\end{abstract}

\section{Introduction}
One of the biggest challenges for deploying autonomous vehicles (AVs) in real life is the requirement of the AVs' capability to interact with surrounding road users. Classifying diverse scenarios and separately designing appropriate decisions using on-hand prior knowledge is unfortunately not realistic \cite{ding2019multi,lin2019intelligent,zhang2019learning,lin2018moha} because of the diversity of scenarios that are far larger and messier than human beings can cope with. A driving encounter is referred to as a scenario where two or multiple vehicles are spatially close to and interact with each other when driving \cite{wang2018clustering}. Intelligent analysis of massive and diverse encounter scenarios is helpful for AVs to make corresponding decisions, for example, at unsignalized intersections \cite{yang2019what}. There are two types of approaches to achieve this: model-based and model-free.
The \emph{model-based} approaches attempt to learn generative models for heterogeneous driving encounters from human driving data \cite{guo2019modeling,ding2019multi}. Alternatively, the proposed \emph{model-free} approach in this work tries to group homogeneous driving encounters in an unsupervised fashion. The advantage of the model-free approach is that there is no need for explicitly modelling underlying complex human driving behaviors.

Driving encounters are essentially multi-vehicle interaction behaviors, which can be described using their trajectories. Therefore, they can be represented as multivariate time series. Time series analysis techniques are popular tools for mining and analyzing trajectories, as reviewed in \cite{feng2016survey}. Unfortunately, most of the existing work focus on single trajectories, rather than multi-vehicle interactive trajectories, see \cite{besse2016review,besse2017destination,choong2017modeling}. For instance, Yao \cite{yao2019clustering} et al. clustered a group of single-driver behaviors such as left turn with multivariate observations (i.e., speed, acceleration, yaw rate, and sideslip angle of the driver) as the clustering index. Dynamic time warping (DTW) techniques were used to measure the similarity between drivers. Niu \cite{niu2019label} et al. developed an efficient clustering algorithm to group several single-vehicle trajectories for road network recognition.

Different from the task of clustering a bunch of single-vehicle trajectories, the fundamental challenges in clustering a bunch of  multi-vehicle trajectories is how to measure the similarity between interactive behaviors. In this paper, we first propose a distance metric for measuring the similarity between two driving encounters. Then, we further formulate and solve a top-$k$ query problem -- \emph{``Given a driving encounter, are we able to find the $k$ most similar driving encounters?"}. We also develop classification and clustering methods based on the similarity. Note that in this paper we mainly emphasize the developed time-series similarity measure and validate it through investigating pair-wise interactive driving behaviors at intersections, since one driving encounter is a multivariate time series encoding all information on two vehicles' trajectories. We apply and extend the matrix profile techniques to efficiently measure similarity between two driving encounters, i.e., two multivariate time series \cite{yeh2016matrix,zhu2016matrix,zhu2018matrix}. Intuitively, two time series are similar if they have substantial similar subsequences. The significance of matrix profile is that it uses a novel representation to efficiently store and compute the nearest neighbour of each subsequence in two time series. Unfortunately, the existing matrix profile work focuses more on univariate time-series data. A new related work \cite{yeh2017matrix} indeed discusses the problem of discovering motifs in multivariate time series. However, it aims at finding patterns in one single multivariate sequence, rather than measuring similarity between a pair of \emph{multivariate and unequal-length} time series.

This work makes the following contributions:

\begin{enumerate}
    \item It extends the matrix profile techniques to solve the pair-wise multivariate time series similarity measure problem;
    \item It proposes an online variant of the multivariate matrix profile method for streaming data;
    \item It is applied to find similar interactive driving behaviors in real-world traffic environments.
\end{enumerate}

The rest of the paper is organized as follows. Section \ref{sec:background} introduces preliminary definitions as background materials. Section \ref{sec:dataset} introduces data description and data pre-processing. Section \ref{sec:algorithm} details our developed new algorithms. Section \ref{sec:experiment} shows the experimental results, followed by the conclusive remarks and future work in Section \ref{sec:conclusion}.

\section{Preliminaries and Definitions}
\label{sec:background}
\begin{definition}{\emph{Time series}:}
A multivariate time series $\bm{T}\in \mathbb{R}^{n \times d}$ is a sequence of real-valued vectors $\bm{t}_i$: $\bm{T} = [\bm{t}_1, \bm{t}_2, \dots, \bm{t}_n]$, where $n$ is the length of $\bm{T}$, and $d$ is the dimension of variables in each vector. A univariate time series is represented by a non-bold symbol $T$.
\end{definition}

\begin{definition}{\emph{Subsequence}:}
A subsequence $\bm{T}_{i, l}$ (or simplified as $\bm{T}_{i}$) of $\bm{T}$ is a continuous subset of the vectors from $\bm{T}$ of length $l$, starting from position $i$; for example, $\bm{T}_{i} = [\bm{t}_i, \bm{t}_{i+1}, \dots, \bm{t}_{i+l-1}]$, where $1 \le i \le n-l+1$. A univariate subsequence is represented by $T_{i}$.
\end{definition}

The first advantageous feature of matrix profile is its light-weight representation for storing two time series' relative distance. The most straightforward way to measure the distance of two time series $\bm{A}$ and $\bm{B}$ is using a $|\bm{A}| \times |\bm{B}|$ size matrix, where the element $dist(\bm{A}_i, \bm{B}_j)$ in the matrix is the distance between the subsequence $\bm{A}_i$ from time series $\bm{A}$ and the subsequence $\bm{B}_j$ from time series $\bm{B}$. Note that $|\bm{A}|$ and $|\bm{B}|$ are the number of subsequences of $\bm{A}$ and $\bm{B}$, respectively. Such a method of storage has a low space efficiency for long time series: even taking into account the symmetry of the matrix, the size is only reduced by half. The second significance of matrix profile is {its high efficiency of computing the distance between two subsequences by using a fast Fourier transform (FFT)}. In the following, we will introduce the notations of matrix profile, matrix profile index, and their learning algorithms in detail.

\begin{definition}\emph{Similarity join set:}
Given all subsequence of $\bm{A}$ and $\bm{B}$, a similarity join set $\bm{J_{AB}}$ for $\bm{A}$ and $\bm{B}$ is a set containing pairs of each subsequence in $\bm{A}$ and its nearest neighbor in $\bm{B}$: $\bm{J_{AB}} = \{\left< \bm{A}_i, \bm{B}_j \right> | \theta_{1nn} (\bm{A}_i, \bm{B}_j)\}$, denoted as $\bm{J_{AB}} = \bm{A} \bowtie \bm{B}$, where $\theta_{1nn} (\bm{A}_i, \bm{B}_j)\}$ is Boolean function which is TRUE if $\bm{B}_j$ is the nearest neighbor of $\bm{A}_i$.
\end{definition}

\begin{definition}{\emph{Matrix profile}:} A matrix profile $\bm{P_{AB}}$ is a vector of the Euclidean distance, where each value of the vector is the distance of each pair in $\bm{J_{AB}}$.
\end{definition}

\begin{definition}{\emph{Matrix profile index}:} A matrix profile index $\bm{I_{AB}}$ of a similarity join set $\bm{J_{AB}}$ is a vector of integers, where $\bm{I_{AB_i}} = j$ if $\bm{A}_i,\bm{B}_j \in \bm{J_{AB}}$. The matrix profile index stores the information that for $\bm{A}_i$ the index of its nearest neighbor in $\bm{B}$ is $j$.
\end{definition}

\begin{figure}[t]
  \centering
  \includegraphics[width = 1.0\linewidth]{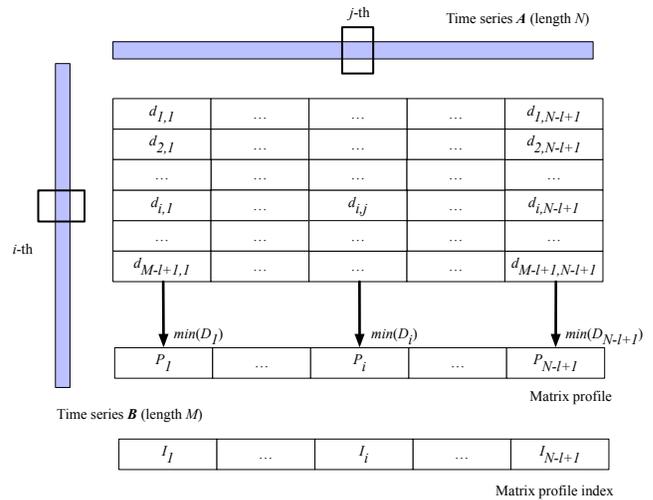}
  \caption{Matrix profile of two time series. \label{fig:matrix_profile}}
\end{figure}

Fig. \ref{fig:matrix_profile} illustrates how to compute the matrix profile and the matrix profile index. The $i$-th row of the matrix is the distance between $i$-th sliding window (subsequence) of $\bm{B}$ (length $M$) over all subsequences of $\bm{A}$ (length $N$). Every row is called a \emph{distance profile}. 

Each element in the matrix profile (the second row from the bottom in Fig. \ref{fig:matrix_profile}) is the minimum value of the corresponding column. The matrix profile index (the bottom row in Fig. \ref{fig:matrix_profile}) vector stores the indexes of these minimum values. Note that the length of the matrix profile and the matrix profile index are both $N-l+1$, where $l$ is the length of each subsequence.

\section{Dataset Description and Preprocessing}
\label{sec:dataset}
The dataset used in this work is the released part of the datasets collected under the University of Michigan Safety Pilot Model Development (SPMD) program \cite{bezzina2015safety,wang2017much}. The data were collected at a sampling frequency of 10 Hz. The dataset for the multi-vehicle interaction analysis used in this work is from the dedicated short-range communication (DSRC) devices mounted on the vehicles. One driving encounter in this paper is essentially the trajectories from two spatially close vehicles, i.e., the GPS devices start to record a pair of interacting vehicles' trajectories when their relative distance is smaller than roughly 100 meters. One driving encounter is a time series of size $n \times 6$, where the features are speed of vehicle \#1, latitude of vehicle \#1, longitude of vehicle \#1, speed of vehicle \#2, latitude of vehicle \#2, longitude of vehicle \#2. The data are collected from three intersections of Michigan City. 


Note that the variation of the longitude and the latitude values at an intersection is small. To better analyze the behaviors, we amplify the position information by transferring the longitude and the latitude into relative distance (unit: meter) from reference positions. In each intersection, the reference position is the spot with the minimum longitude value and the minimum latitude value among all trajectories in the dataset. As a result, the original features are transformed into the speed of vehicle \#1 ($v_1$), the relative latitude distance of vehicle \#1 ($y_1$), the relative longitude distance of vehicle \#1 ($x_1$), the speed of vehicle \#2 ($v_2$), the relative latitude distance of vehicle \#2 ($y_2$), and the relative longitude distance of vehicle \#2 ($x_2$).

\section{proposed Methodology}
\label{sec:algorithm}
Algorithm \ref{algo:MMASS} shows how to compute a distance profile given a time series subsequence and another complete time series, i.e., how to get a row as shown in Fig. \ref{fig:matrix_profile}.
Instead of directly computing the Euclidean distance, the algorithm first conducts a $sliding\_dot\_product$ operation to obtain the inner products using an FFT (line 3). Second, the mean and the standard deviation of the subsequence and the time series are computed (line 4). Last, computing the Euclidean distance from the dot product by $compute\_distance\_profile$ uses the following formula (line 5):

\begin{equation}
\label{equa:distance_euclidean}
    D_j^i=\sqrt{2 l\left(1-\frac{QT^i_j-l \mu_{Q}^i {M_T^i}_j}{l \sigma_{Q}^i {\Sigma_T^i}_j}\right)}
\end{equation}
The line 6 inside the \emph{for} loop states that the algorithm computes a distance profile for each dimension and combines them into a multivariate distance profile.
The efficiency is significantly improved compared with the brute-force computation of the Euclidean distance. The details will be discussed in the following algorithms and complexity analysis subsections.

\begin{algorithm}[t]
\LinesNumbered
\caption{Multivariate distance profile computation:\label{algo:MMASS}}
\KwIn{A multivariate query $\bm{Q}$ (e.g., one subsequence of time series $\bm{B}$ in Fig. \ref{fig:matrix}) and a time series $\bm{T}$ (e.g., time series A in Fig. \ref{fig:matrix})}
\KwOut{A distance profile $\bm{D}$ of $\bm{Q}$ over $\bm{T}$}
Initialize $\bm{D}$ as a zero vector

\For{all dimension $d$}
{
$\bm{QT}^i := sliding\_dot\_product(Q^i, T^i)$;

$\mu_Q^i, \sigma_Q^i, \bm{M}_T^i, \bm{\Sigma}_T^i := compute\_mean\_std(Q^i, T^i)$;

$\bm{D}^i = compute\_distance\_profile(Q^i, T^i, \bm{QT}^i,\\ \mu_Q^i, \sigma_Q^i, \bm{M}_T^i, \bm{\Sigma}_{T}^i)$;

$\bm{D} = \bm{D} + \bm{D^i}$

}
$\bm{D} = \frac{\bm{D}}{d}$

Return $\bm{D}$
\end{algorithm}

\subsection{Multivariate STAMP}
In this paper, we extend the \emph{STAMP} (Scalable Time series Anytime Matrix Profile) algorithm \cite{yeh2016matrix} to the multivariate version, hereafter referred to as \emph{MUSTAMP}. MUSTAMP randomly selects one row of the matrix (as shown in Fig. \ref{fig:matrix_profile}) in each iteration, computes the corresponding distance profile, and updates the minimum values and the index into the matrix profile and the matrix profile index. Note that the matrix as shown in Fig. \ref{fig:matrix_profile} is only for illustration purposes. It is not necessary to store such a large matrix, because MUSTAMP can continuously maintain and update the minimum values and index in every iteration.

\textbf{Complexity analysis for MUSTAMP}: As shown in Equation \ref{equa:distance_euclidean}, computing the mean and the standard deviation can be achieved with $O(1)$ time complexity \cite{rakthanmanon2013addressing}. So the dominant term of the complexity is from $sliding\_dot\_product$, of which the complexity is $O(n\log n)$ instead of $O(nl)$ for a brute-force solution, where $n$ is the length of the time series (or the longest length of two time series $\bm{A}$ and $\bm{B}$). The difference is more significant for a larger subsequence length $l$. Since we need to obtain $n$ multidimensional distance profiles (i.e., $n$ rows in Fig. \ref{fig:matrix_profile}), the overall complexity is $O(d n^2 \log n)$. Ref. \cite{yeh2016matrix} shows that the empirical runtime for computing a distance profile is roughly $O(n)$ instead of $O(n\log n)$ owing to the well-optimized FFT in many programming platforms, therefore the overall empirical runtime would be roughly $O(d n^2)$.

\subsection{Multivariate STOMP}
\begin{figure}[t]
      \centering
      \includegraphics[width=0.45\textwidth]{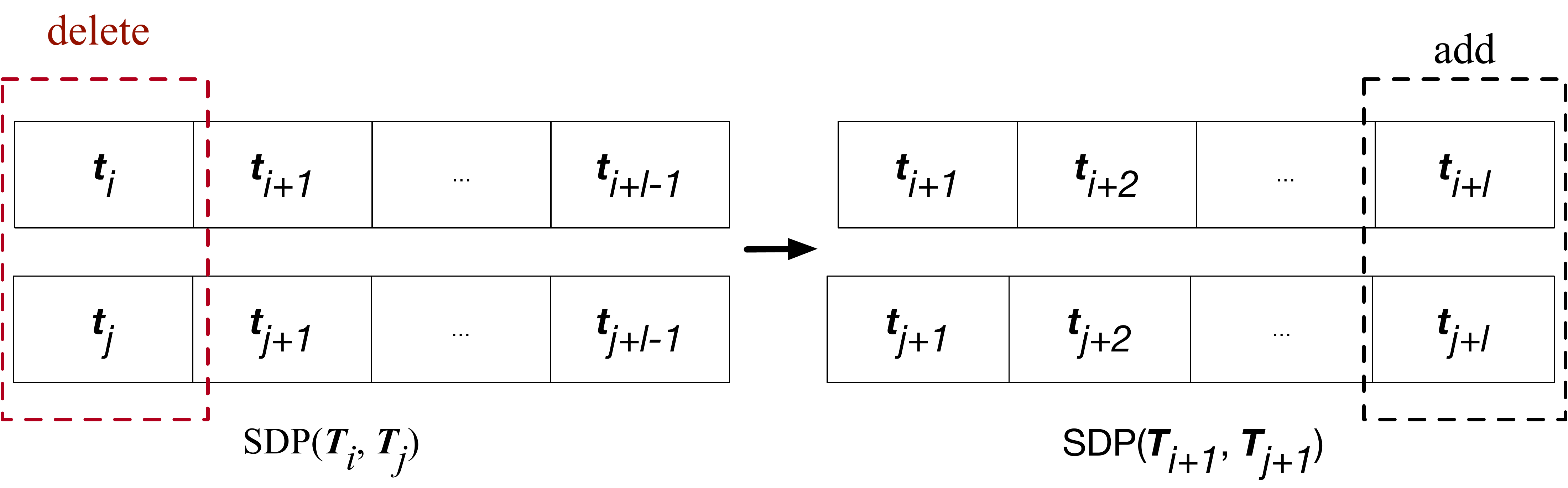}
      \caption{The relation of two adjacent dot products of MUSTOMP\label{fig:stomp}}
      \label{fig:matrix}
   \end{figure}

As discussed before, the STAMP algorithm randomly selects a row in the matrix shown in Fig. \ref{fig:matrix_profile} in each iteration. In addition, the computations of different rows are independent. Ref. \cite{zhu2016matrix} discovers that actually any two adjacent rows' distance profiles have a relation. We extend the results and implement a multivariate version of STOMP, hereafter referred to as \emph{MUSTOMP}. The key idea is that once we operate a $sliding\_dot\_product$ between two multivariate subsequences to obtain $SDP(\bm{T}_i, \bm{T}_j)$, we can incrementally compute $SDP(\bm{T}_{i+1}, \bm{T}_{j+1})$ using the following formula:

\begin{multline}
    SDP(\bm{T}_{i+1}, \bm{T}_{j+1})=SDP(\bm{T}_i, \bm{T}_j) - \sum\limits_{k=1}^{d} t_{i}^{k}t_{j}^{k}\\
    +\sum_{k=1}^{d}t_{i+l-1}^{k}t_{j+l-1}^{k}
\end{multline}
Fig. \ref{fig:stomp} illustrates that realization from $SDP(\bm{T}_i, \bm{T}_j)$ to $SDP(\bm{T}_{i+1}, \bm{T}_{j+1})$ just needs to remove the product ${\bm{t}_i\bm{t}_j}$ (the red block, i.e. $\sum_{k=1}^{m} t_{i}^{k}t_{j}^{k}$) and add the product $\bm{t}_{i+l-1}\bm{t}_{j+l-1}$ (the black block, i.e., $\sum_{k=1}^{m}t_{i+l-1}^{k}t_{j+l-1}^{k}$). The remaining products are kept without the need of recomputing.
   
\textbf{Complexity analysis of MUSTOMP}: According to the relation between $SDP(\bm{T}_i, \bm{T}_j)$ and $SDP(\bm{T}_{i+1}, \bm{T}_{j+1})$, for the one-dimensional case, we can obtain the distance profile in $O(n)$ time complexity instead of STAMP's $O(n\log n)$. For the multivariate case, the overall complexity is $O(dn^2)$.

\subsection{Online Learning Algorithm of MUSTAMP}
The MUSTAMP and MUSTOMP algorithms are essentially both for batch learning. Batch learning means that we need to see the entire time series $\bm{A}$ and $\bm{B}$ before computing the matrix profile. However, in practice, we usually only see the incomplete interactive behaviors, i.e., a part of the time series. It becomes crucial to have an online incremental variant of the algorithm, which is capable of updating the matrix profile by taking care of the new arriving data points rather than restarting from scratch.

\begin{figure}[t]
      \centering
      \includegraphics[width=0.48\textwidth]{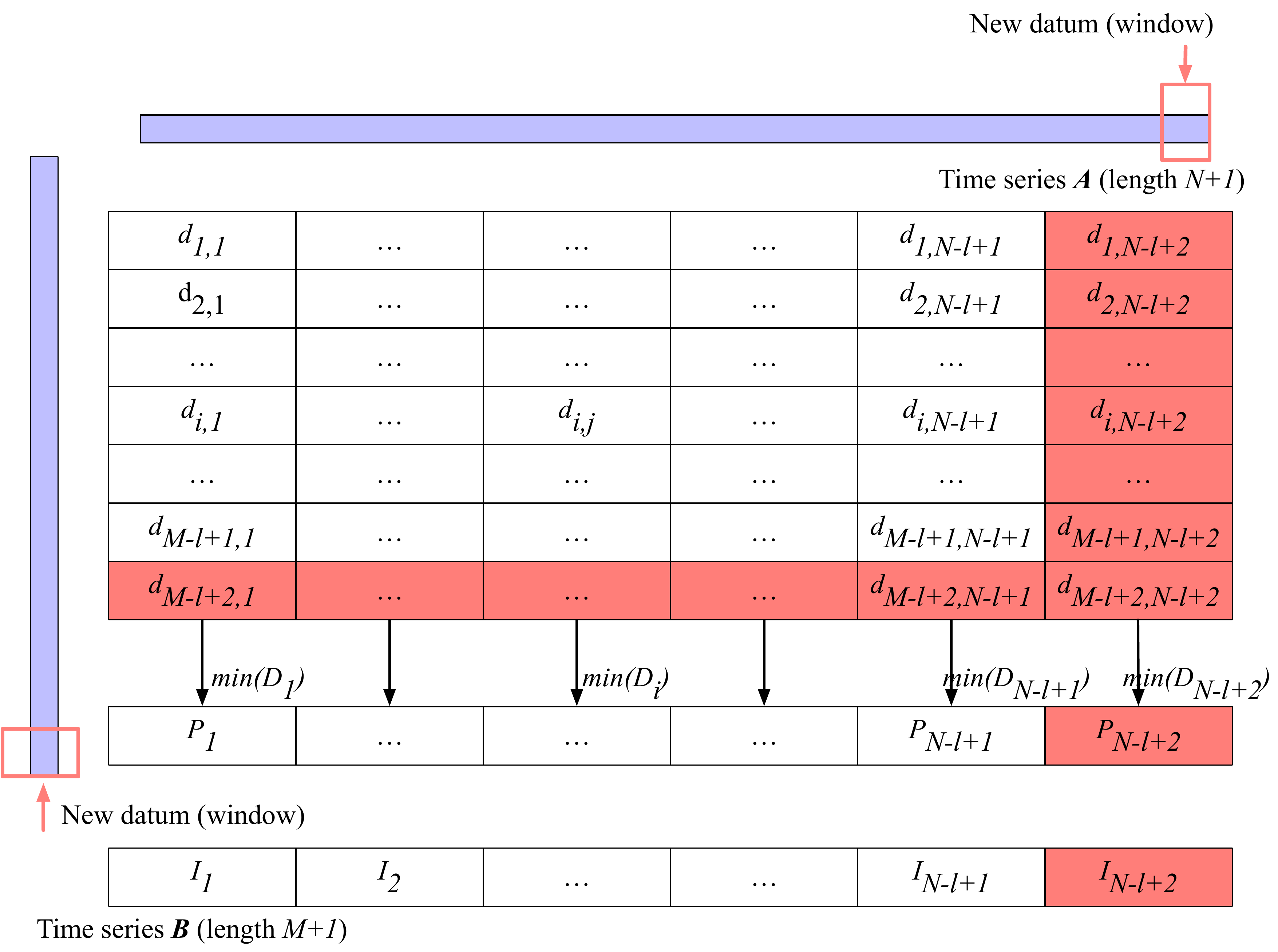}
      \caption{Online maintenance of matrix profile. Compared with the matrix shown in Fig. \ref{fig:matrix}, the size of the distance profiles in the rows and the columns is growing with the newly arrived data. \label{fig:matrix_online}}
   \end{figure}

Fig. \ref{fig:matrix_online} illustrates the idea of the incremental maintenance for a matrix profile. Compared to the matrix profile we already have (see Fig. \ref{fig:matrix_profile}) for the data we have seen so far, we have two new data points of time series $\bm{A}$ and $\bm{B}$, i.e., two sliding windows. The red row $d_{M-l+2, 1} \cdots d_{M-l+2, N-l+2}$ is the new distance profile about the new sliding window of time series $\bm{B}$ over time series $\bm{A}$. Similarly, the new red column is the new distance profile about the new sliding window of time series A over time series B. The matrix profile $P_1 \cdots P_{N-l+1}$ and the matrix index $I_1 \cdots I_{N-l+1}$ both get updated since we have the new vector $d_{N-l+2, 1} \cdots d_{N-l+2, N-l+1}$. We also need to append new items $P_{N-l+2}$ and $I_{N-l+2}$ according to the minimum value and its index of the new column.

\subsection{Distance measure for top-k query}
The matrix profile is a vector storing the subsequential similarity ``fingerprint" of two time series. However, normally we need a scalar value to represent the distance between two time series. In the following, we will first introduce how to ``compress" the matrix profile from a vector to a scalar. Then, the distance metric is plugged into a top-$k$ query algorithm, see Algorithm \ref{algo:top-k}. The top-$k$ query task is: given a query time series, how to retrieve the $k$ most similar time series from a dataset. It is widely used to validate the robustness of a similarity metric in the time series data mining area \cite{bu2007wat,grabocka2014learning}.
\begin{algorithm}[h]
\setcounter{AlgoLine}{0}
\LinesNumbered
\caption{top-$k$ query algorithm:\label{algo:top-k}}
\KwIn{All time series in the dataset (number of time series: N), threshold $thr$}
\KwOut{top-$k$ most self-exclusively similar time series for each time series}

\For{i in range($1, N$)}
{
\For{j in range($i+1, N$)}
{
$linkage(i, j) = MUSTA/OMP (\bm{T}^{i}, \bm{T}^{j})$; // {MUSTA/OMP means MUSTAMP or MUSTOMP}

$distance(i, j) = 1 - \frac{2 \times \text{\# of elements in linkage}(i, j)<=thr}{|\bm{T}^{i}|+|\bm{T}^{j}|}$;

$distance(j, i)=distance(i, j)$;

}
sorted $distance(i, \cdot)$, and select $k$ self-exclusive time series with the smallest distance values.


}

Return top-$k$ query for each time series
\end{algorithm}

The algorithm first sets up the linkages for the query time series $\bm{T}[i]$ (a complete sequence) and other complete time series (complete sequences). Note that to save the computation time, we only compute the distance forwards: $\bm{T}[1]$ with $\bm{T}[2]$ until $\bm{T}[N]$; then $\bm{T}[2]$ with $\bm{T}[3]$ until $\bm{T}[N]$. $distance(\bm{T}[2], \bm{T}[1])$ is obviously equal to $distance(\bm{T}[1], \bm{T}[2])$ without another redundant computation. Line 4 is the formula for the transformation from the matrix profile to the scalar distance value. Intuitively, two time series are similar if the values of the matrix profile are small, i.e., for any subsequence $\bm{A}$, its relative distance to the nearest neighboring subsequence in $\bm{B}$ is always small. Namely, they have substantial similar subsequences. The number of elements in $linkage(i, j) <= thr$ in the numerator is the number of significantly similar subsequences. A larger number implies a small distance value. The distance value varies from 0 to 1.

One disadvantage of the original matrix profile algorithm is the asymmetry problem, i.e.,  $\mathbf{J_{AB}} \neq \mathbf{J_{BA}}$ and $\mathbf{P_{AB}} \neq \mathbf{P_{BA}}$. The asymmerty property will be problematic when we compute the distance between the time series $\bm{T}[i]$ and $\bm{T}[j]$, because intuitively we want $distance(i, j) = distance(j, i)$. To make a deterministic computation, we deal with the problem by checking the following different conditions:

\begin{enumerate}
    \item when $|\bm{T}[i]| < |\bm{T}[j]|$, $linkage(i, j) = MUSTA/OMP(\bm{T}[i], \bm{T}[j])$;
    \item when $|\bm{T}[j]| < |\bm{T}[i]|$, $linkage(i, j) = MUSTA/OMP(\bm{T}[j], \bm{T}[i])$;
    \item when $|\bm{T}[j]| == |\bm{T}[i]|$, we compute $MUSTA/OMP(\bm{T}[i], \bm{T}[j])$, $MUSTA/OMP(\bm{T}[j], \bm{T}[i])$, and their corresponding distance. The final $distance(i,j)$ is the average value.
\end{enumerate}
Line 7 in Algorithm \ref{algo:top-k} sorts all neighborhood time series and selects the top-$k$ nearest neighborhoods.

\section{Experimental Results}
\label{sec:experiment}
This section discusses the experimental details about how to apply the algorithms to deal with the similarity measure for driving encounters. However, our proposed algorithm is general enough to deal with other interactive behaviors, such as vehicle-pedestrian and vehicle-cyclist interactions.
\subsection{Top-$k$ Driving Encounter Query Using MUSTAMP}
\begin{figure}[t]
\centering
\subfigure[query driving encounter]{
\includegraphics[width=0.24\textwidth]{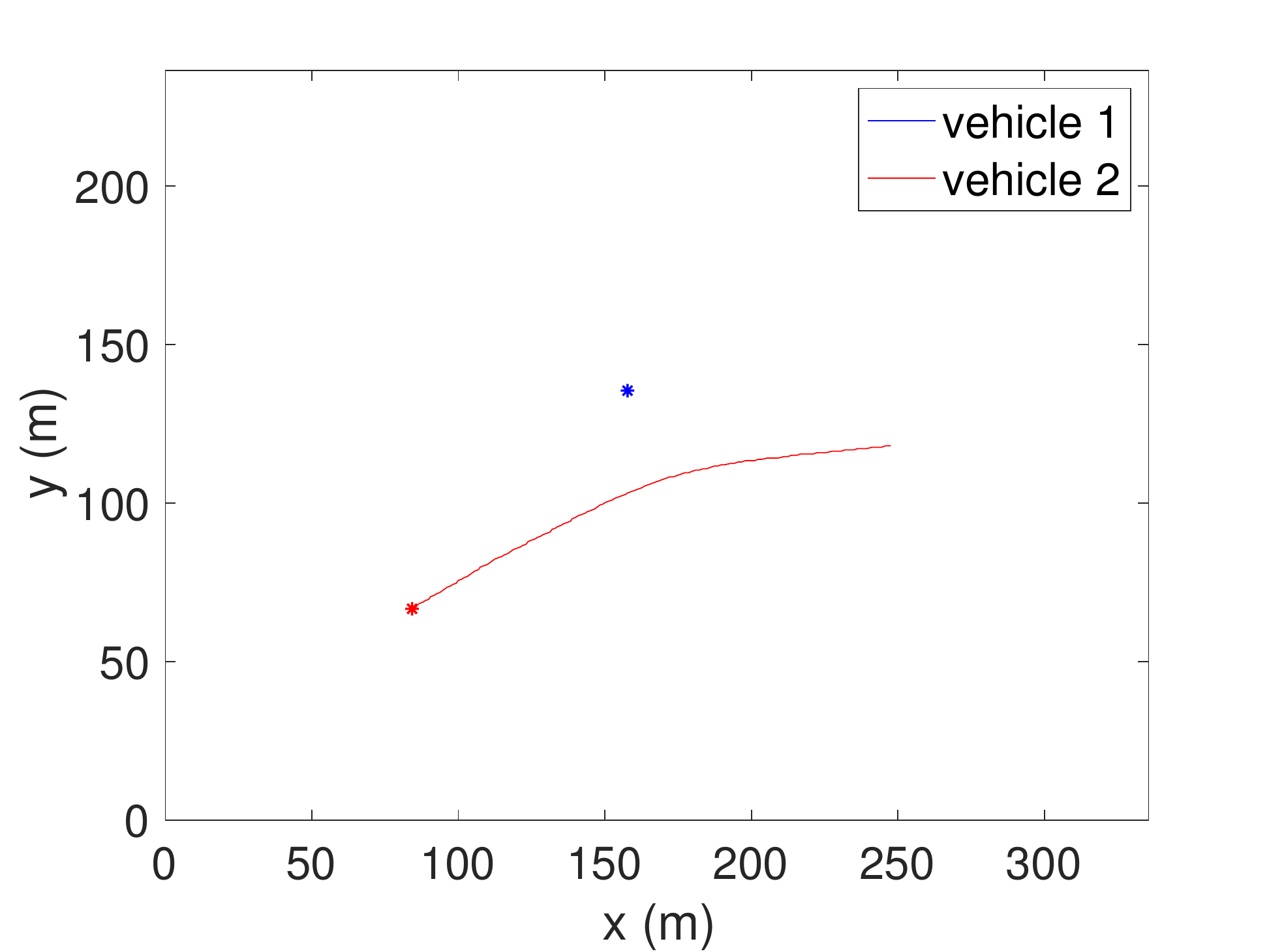}
}%
\subfigure[top-1 most similar encounter]{
\includegraphics[width=0.24\textwidth]{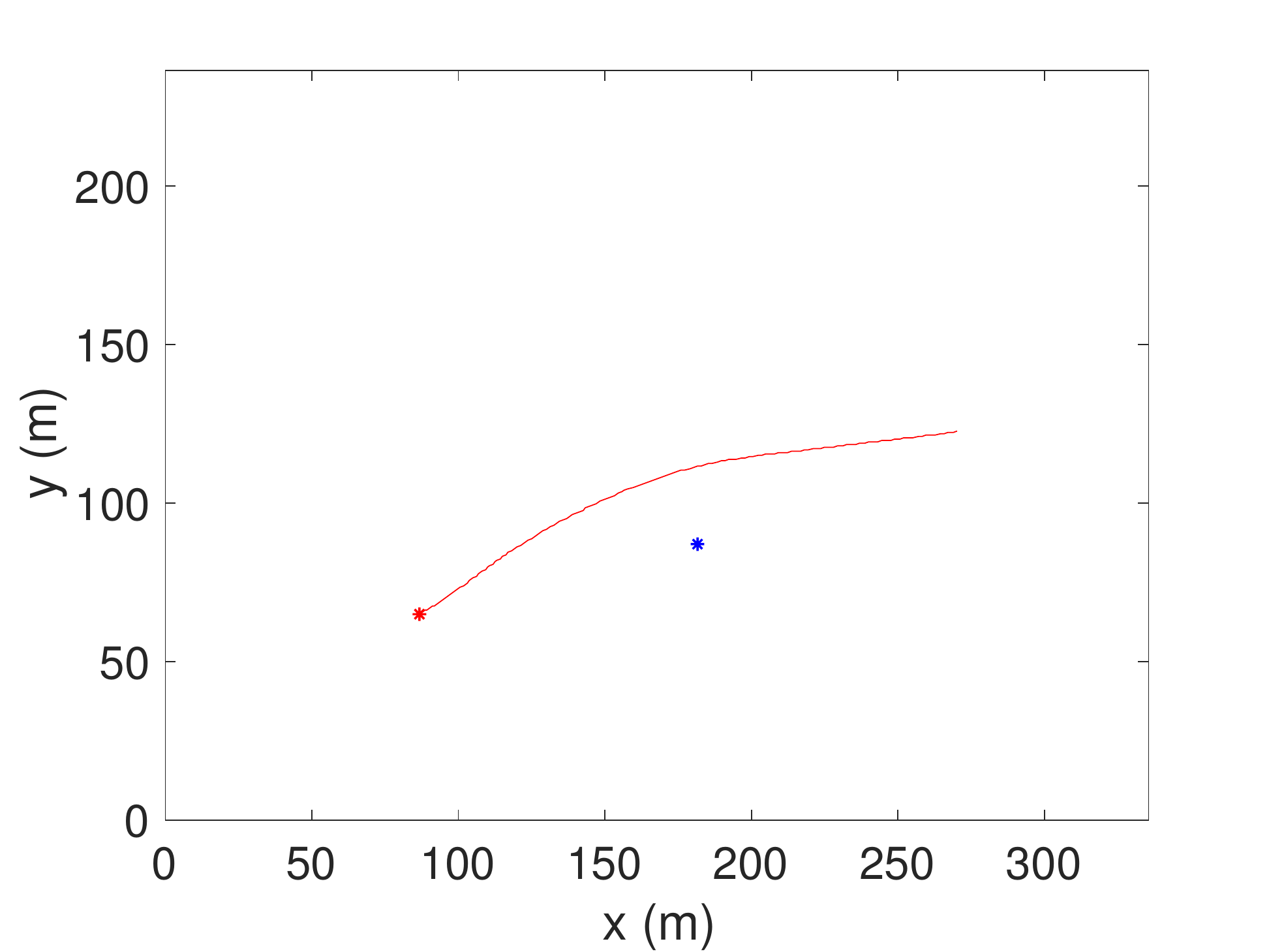}
}\\
\subfigure[top-2 most similar encounter]{
\includegraphics[width=0.24\textwidth]{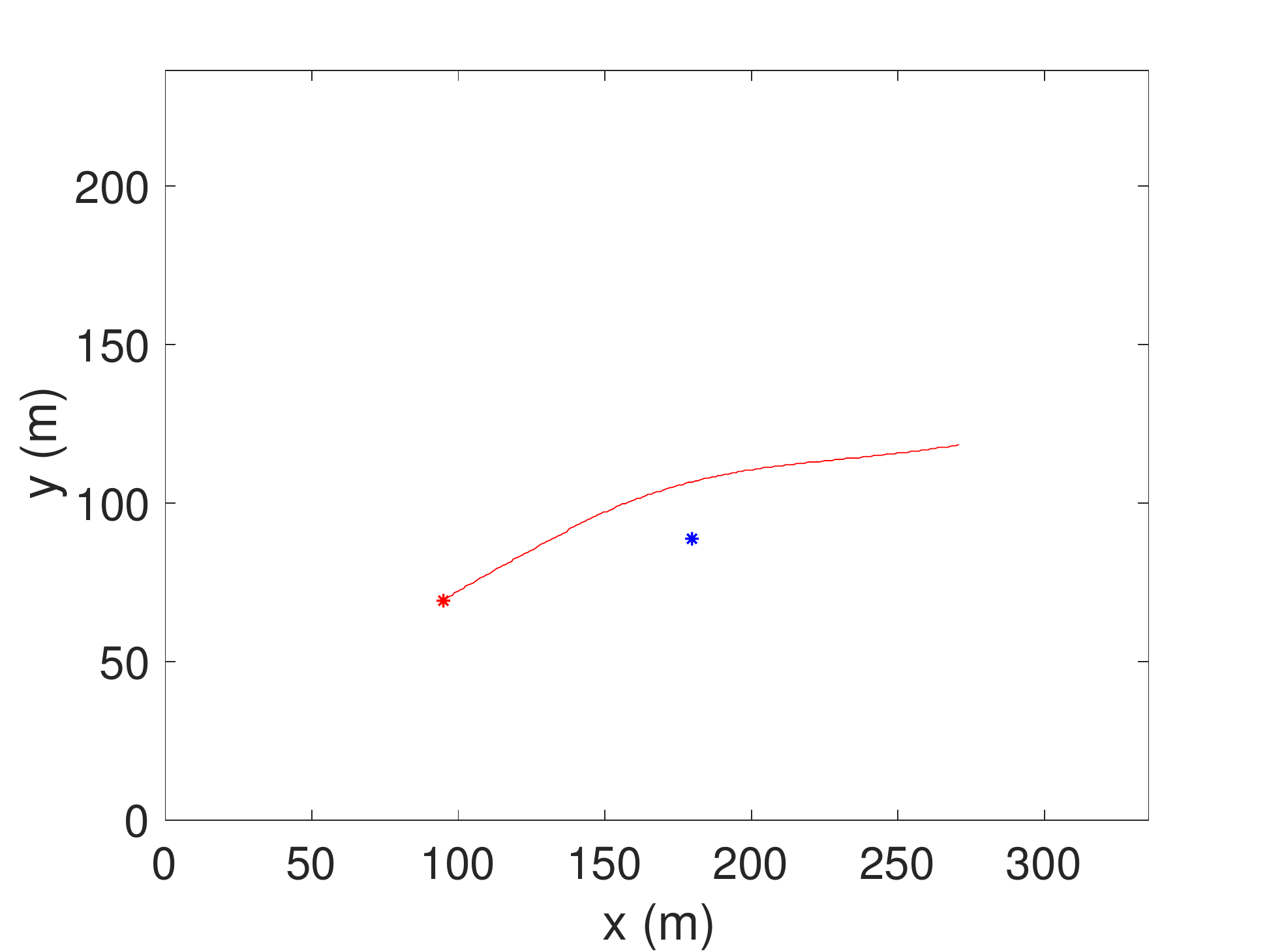}
}%
\subfigure[top-3 most similar encounter]{
\includegraphics[width=0.24\textwidth]{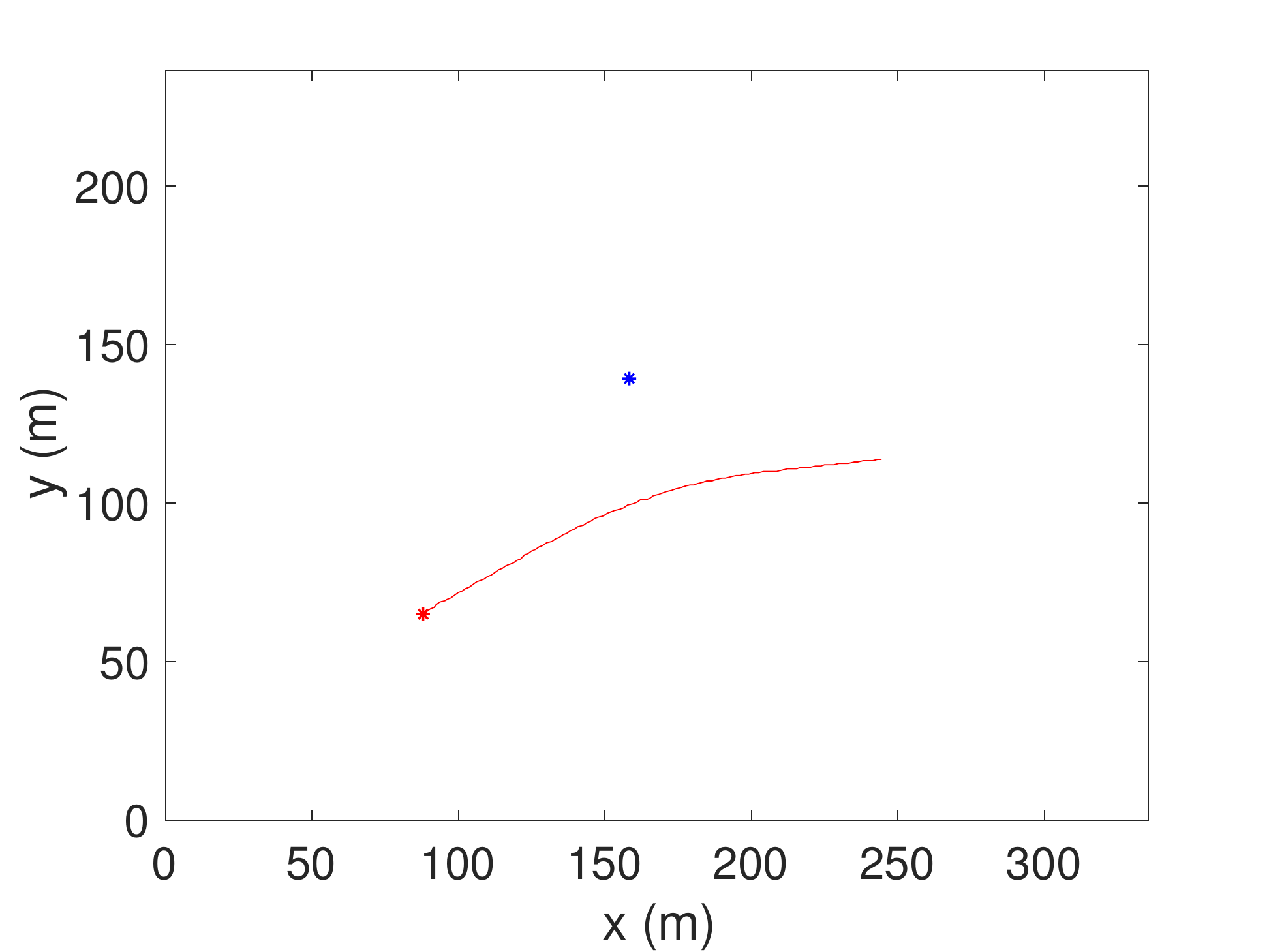}
}%
\caption{An example of the top-$k$ query for finding the top-$k$ most similar encounters. The blue star and the red star indicate the starting positions of the two trajectories. The distance of a\&b, a\&c, a\&d are 0.0728, 0.0763, and 0.1415, respectively\label{fig:top-k}.}
\end{figure}

The subplot (a) in Fig. \ref{fig:top-k} is from a driving encounter in which vehicle 1 is stopping and waiting to proceed while vehicle 2 is moving from the west to the east. Our algorithm finds the top-3 most similar interactive behaviors (b, c, and d) from the dataset. We can observe that owing to the normalization, the algorithm can generalize to the topological minor behaviors, i.e., having vehicle 1 on the left-hand side and the right-hand side of vehicle 2 are recognized as similar behaviors.

\begin{figure}[t]
      \centering
      \includegraphics[width=0.3\textwidth]{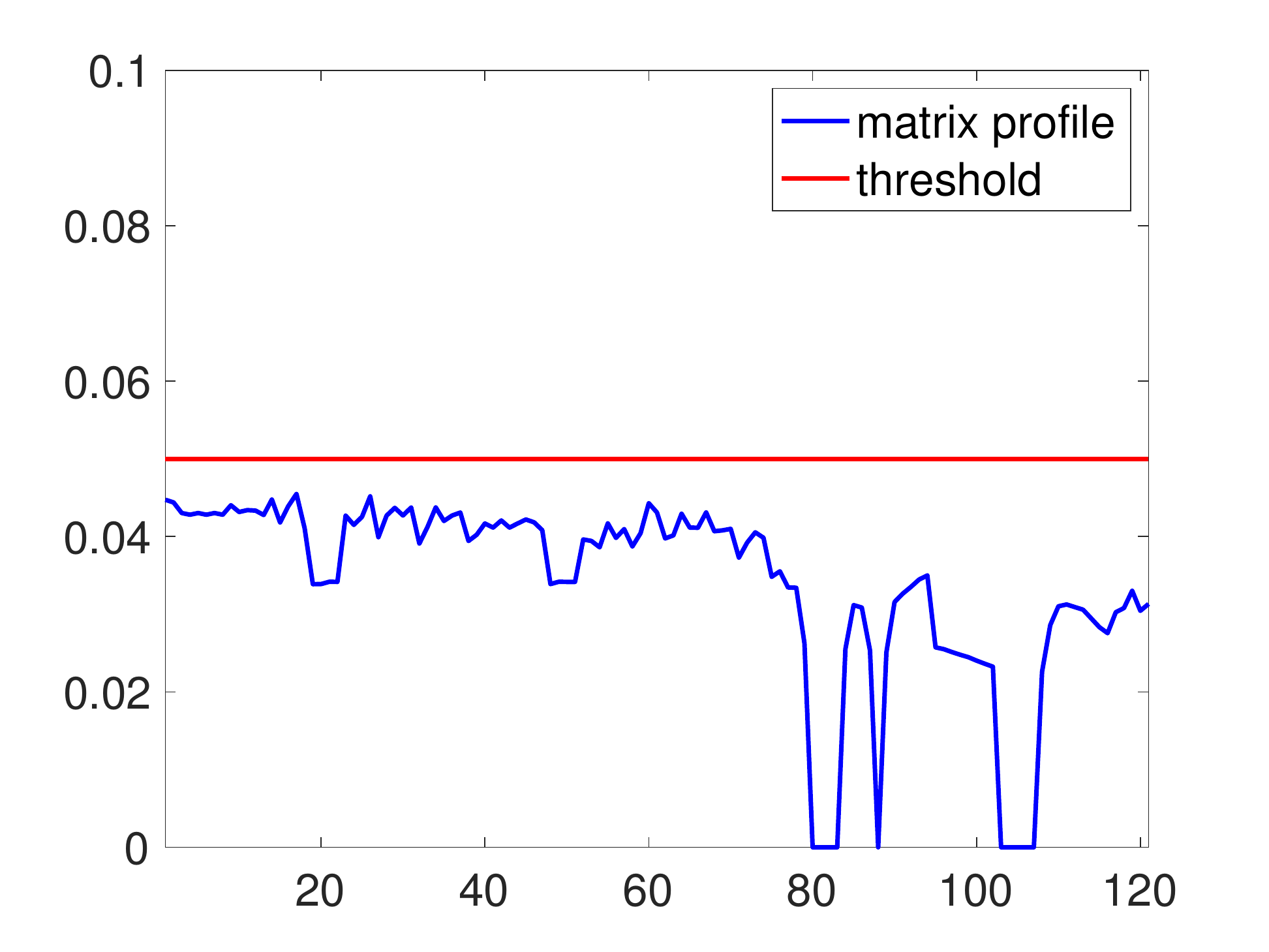}
      \caption{Distance metric example}
      \label{fig:distance_metric}
   \end{figure}

Fig. \ref{fig:distance_metric} shows an example of how to compute the distance value from a matrix profile (line 4 in Algorithm \ref{algo:top-k}) for the encounters (a) and (b). First, we compare the matrix profile with the threshold and get the number in the numerator 2*121=242. The denominator is the sum of the two series' subsequence number 261, and the final distance value is therefore $1-\frac{242}{261}=0.0728$.

\subsection{Online Convergence Evaluation}

\begin{figure}[t]
      \centering
      \includegraphics[width=0.3\textwidth]{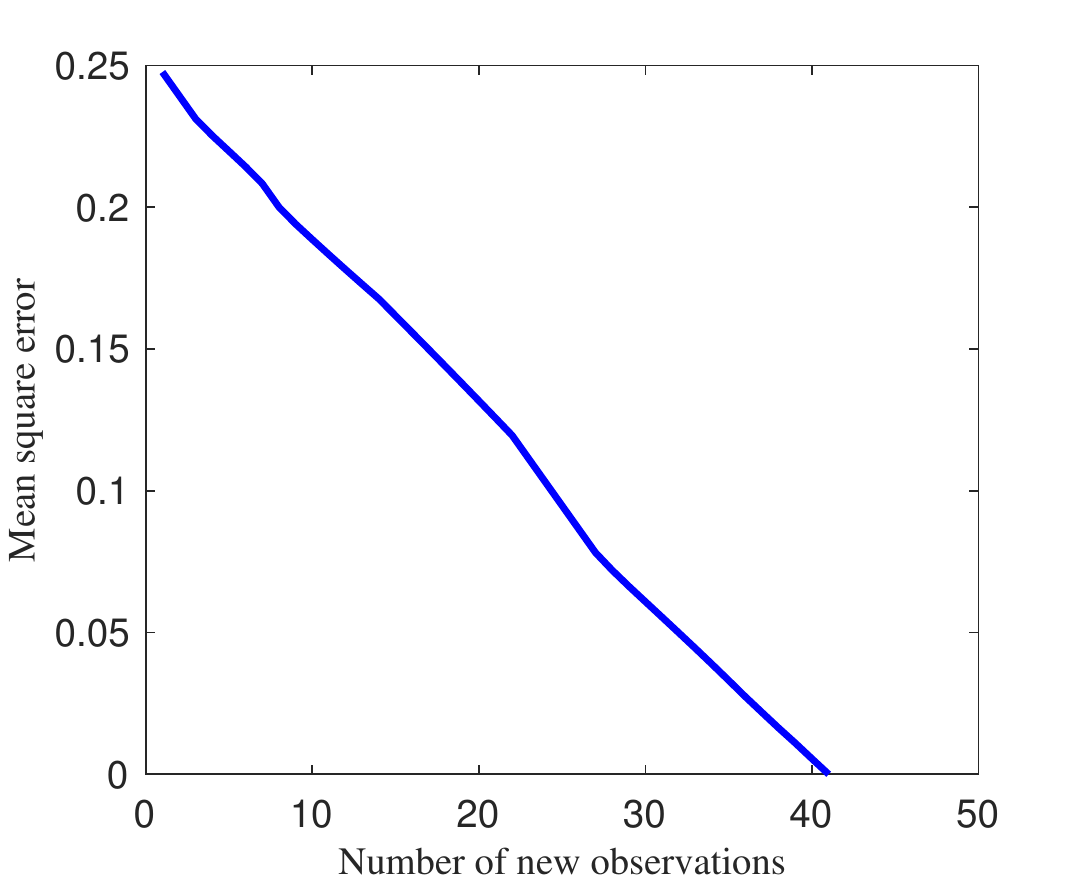}
      \caption{Convergence of incremental maintenance}
      \label{fig:Convergence}
   \end{figure}

The example used for evaluating the online version of MUSTAMP is again from the two encounters (a) and (b) in Fig. \ref{fig:distance_metric}. The length of $\bm{A}$ and $\bm{B}$ are 140 and 159, respectively. We first calculate the matrix profile from $\bm{A}(1:100)$ and $\bm{B}(1:119)$ by assuming that they have been observed, then using the online learning algorithm to compute the remaining 40 data points. Intuitively, the matrix profile will converge to the batch learning for $\bm{A}(1:140)$ and $\bm{B}(1:159)$. Fig. \ref{fig:Convergence} shows the convergence result. The y-axis value is the mean square error between the current matrix profile and the matrix profile of the complete two time series.

\subsection{Classification and Clustering}
\begin{figure}[t]
      \centering
      \includegraphics[width=0.5\textwidth]{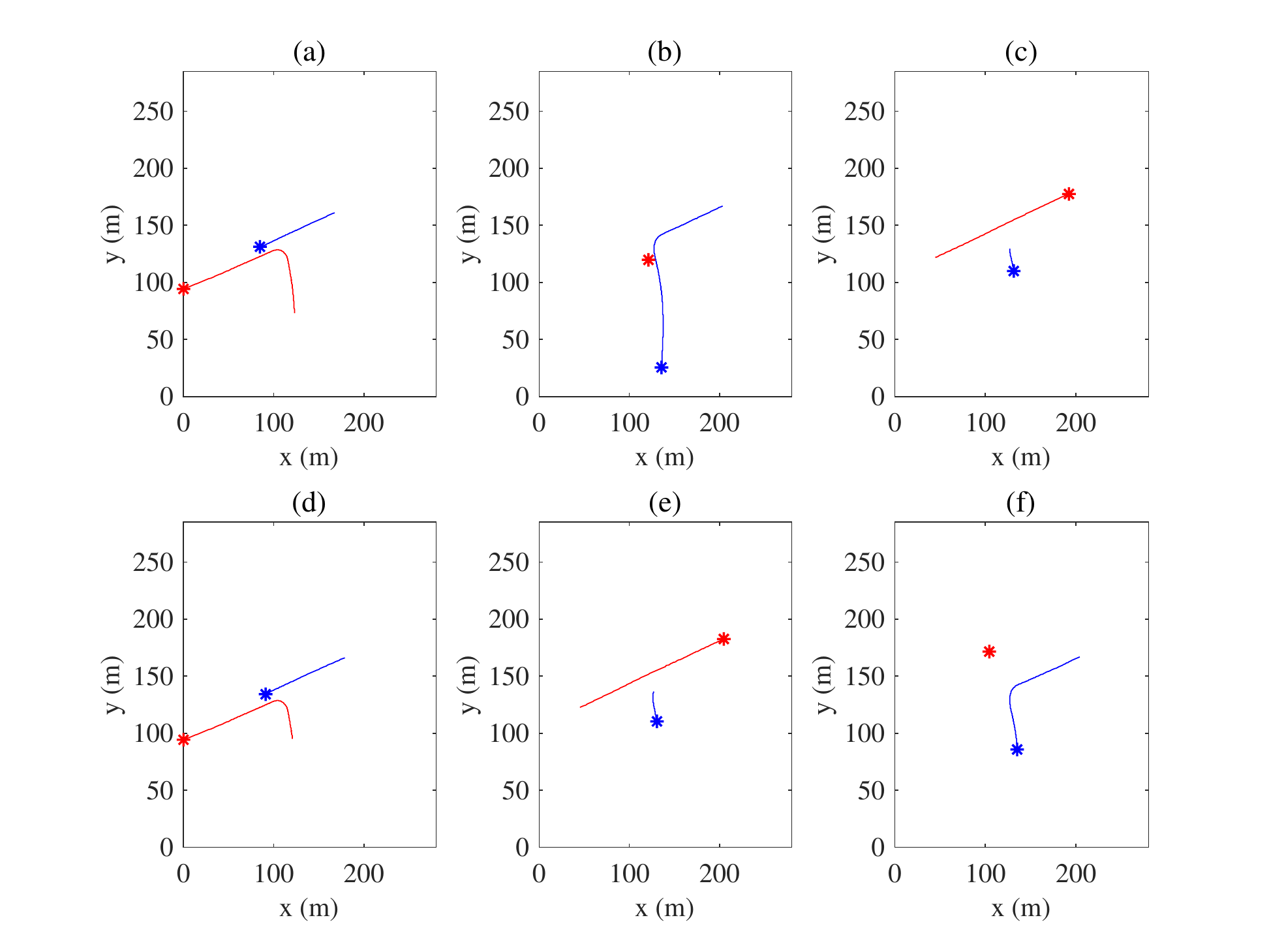}
      \caption{Classification of new observations. The distance values between (d) and (a-c) are 0.05, 1, 1; the distance values between (e) and (a-c) are 1, 1, 0.0950; the distance values between (f) and (a-c) are 1, 0.1863, 1.}
      \label{fig:classification}
   \end{figure}

Classification is a classical task of machine learning for identifying to which category a new observation belongs. For AVs, we might have some predefined scenarios as semantically labelled categories. AVs are able to match the new observations to the specified categories and take the corresponding decisions. Fig. \ref{fig:classification} shows an example of classification based on the  proposed distance metric. The first column is from three frequently seen behaviors in the dataset. Using human interpretation: the (a) scenario has two cars driving in the same direction at the beginning, but then the red vehicle turns right at the intersection; in the (b) scenario, the red car stops while the blue car approaches the red car and then turns right; in the (c) scenario, the blue car approaches the intersection while the red car goes straight. The new observations in the second row are assigned to the most similar behavior in the first row.

In many cases of practice, we do not know the number of categories in advance, since human labelling is costly. Hierarchical clustering is an unsupervised learning approach to automatically obtain categories according to the similarity metric. The startpoint is that every driving encounter is a cluster, and the algorithm continuously merges the two most similar clusters into a new cluster in each iteration. The algorithm terminates according to some stopping criterion such as number of clusters or distance threshold.


\subsection{Run-time Comparison}
Table \ref{tab:runtime} is the summary of run-times for each intersection dataset. The $n$ driving encounters would require that the number of pairs we need to compute the distance is their permutations $C^2_{n}$. The average computation cost for one pair is less than 200 ms. The average computation cost of every iteration for the online MUSTAMP is around 2 ms, which is able to handle the streaming traffic data because the sampling frequency of the data used in this work is 10 Hz, i.e., the new datum arrives every 100 ms. All the computations are done on a standard Macpro laptop with an Intel 2.5 GHz i5 core and 8 GB RAM.
\begin{table}[ht]
\caption{Run-time of MUSTAMP\label{tab:runtime}}
\centering
\resizebox{.47\textwidth}{!}{
\begin{tabular}{cccc}
\hline
 Intersection & $C^2_{n}$ & MUSTAMP Runtime (s) & MUSTOMP Run-time (s)\\
 \hline
 1& 4851 &  840.1 & 191.4\\
 2& 14878 &  1828.6 & 633.5\\
 3& 20503 & 2618.3 & 920.4\\
 \hline
\end{tabular}}
\end{table}

\section{Conclusion}
\label{sec:conclusion}
In this paper, we propose a novel distance metric to measure similarity between two unequal-length multivariate time series. It has been applied to find similar interactive behaviors of driving encounters in the top-$k$ query, the classification, and the clustering tasks. We also developed an incremental learning variant to handle streaming traffic data. The framework is general for dealing with other on-road interactive behaviors such as vehicle-pedestrian and vehicle-cyclist interactions. In the near future, we will develop a parallel computation version of our approach to make it more applicable in real life. The idea is that, for instance, because the distance profile in MUSTAMP is independent, multi-core processors can be leveraged to independently compute and to aggregate all distance profiles to obtain the final matrix profile.





\section{ACKNOWLEDGMENT}
This material is based upon work supported by the United States Air Force and DARPA under Contract No. FA8750-18-C-0092. Any opinions, findings and conclusions or recommendations expressed in this material are those of the author(s) and do not necessarily reflect the views of the United States Air Force and DARPA.

\bibliographystyle{IEEEconfbibstyle}
\bibliography{root}

\begin{thebibliography}{10}
\providecommand{\url}[1]{#1}
\csname url@rmstyle\endcsname
\providecommand{\newblock}{\relax}
\providecommand{\bibinfo}[2]{#2}
\providecommand\BIBentrySTDinterwordspacing{\spaceskip=0pt\relax}
\providecommand\BIBentryALTinterwordstretchfactor{4}
\providecommand\BIBentryALTinterwordspacing{\spaceskip=\fontdimen2\font plus
\BIBentryALTinterwordstretchfactor\fontdimen3\font minus
  \fontdimen4\font\relax}
\providecommand\BIBforeignlanguage[2]{{%
\expandafter\ifx\csname l@#1\endcsname\relax
\typeout{** WARNING: IEEEtran.bst: No hyphenation pattern has been}%
\typeout{** loaded for the language `#1'. Using the pattern for}%
\typeout{** the default language instead.}%
\else
\language=\csname l@#1\endcsname
\fi
#2}}

\bibitem{ding2019multi}
W.~Ding, W.~Wang, and D.~Zhao, ``A multi-vehicle trajectories generator to
  simulate vehicle-to-vehicle encountering scenarios,'' in \emph{2019
  International Conference on Robotics and Automation (ICRA)}.\hskip 1em plus
  0.5em minus 0.4em\relax IEEE, 2019, pp. 4255--4261.

\bibitem{lin2019intelligent}
Q.~Lin, ``Intelligent control systems: Learning, interpreting, verification,''
  Ph.D. dissertation, Delft University of Technology, 2019.

\bibitem{zhang2019learning}
W.~Zhang and W.~Wang, ``Learning v2v interactive driving patterns at signalized
  intersections,'' \emph{Transportation Research Part C: Emerging
  Technologies}, vol. 108, pp. 151--166, 2019.

\bibitem{lin2018moha}
Q.~Lin, Y.~Zhang, S.~Verwer, and J.~Wang, ``Moha: a multi-mode hybrid automaton
  model for learning car-following behaviors,'' \emph{IEEE Transactions on
  Intelligent Transportation Systems}, vol.~20, no.~2, pp. 790--796, 2018.

\bibitem{wang2018clustering}
W.~Wang, A.~Ramesh, and D.~Zhao, ``Clustering of driving scenarios using
  connected vehicle datasets,'' \emph{arXiv preprint arXiv:1807.08415}, 2018.

\bibitem{yang2019what}
S.~Yang, W.~Wang, Y.~Jiang, S.~Zhang, and W.~Deng, ``What contributes to
  driving behavior prediction at unsignalized intersections?''
  \emph{Transportation Research Part C: Emerging Technologies}, vol. 108, no.
  2019, pp. 100--114, 2019.

\bibitem{guo2019modeling}
Y.~Guo, V.~V. Kalidindi, M.~Arief, W.~Wang, J.~Zhu, H.~Peng, and D.~Zhao,
  ``Modeling multi-vehicle interaction scenarios using gaussian random field,''
  \emph{arXiv preprint arXiv:1906.10307}, 2019.

\bibitem{feng2016survey}
Z.~Feng and Y.~Zhu, ``A survey on trajectory data mining: Techniques and
  applications,'' \emph{IEEE Access}, vol.~4, pp. 2056--2067, 2016.

\bibitem{besse2016review}
P.~C. Besse, B.~Guillouet, J.-M. Loubes, and F.~Royer, ``Review and perspective
  for distance-based clustering of vehicle trajectories,'' \emph{IEEE
  Transactions on Intelligent Transportation Systems}, vol.~17, no.~11, pp.
  3306--3317, 2016.

\bibitem{besse2017destination}
------, ``Destination prediction by trajectory distribution-based model,''
  \emph{IEEE Transactions on Intelligent Transportation Systems}, vol.~19,
  no.~8, pp. 2470--2481, 2017.

\bibitem{choong2017modeling}
M.~Y. Choong, L.~Angeline, R.~K.~Y. Chin, K.~B. Yeo, and K.~T.~K. Teo,
  ``Modeling of vehicle trajectory clustering based on lcss for traffic pattern
  extraction,'' in \emph{2017 IEEE 2nd International Conference on Automatic
  Control and Intelligent Systems (I2CACIS)}.\hskip 1em plus 0.5em minus
  0.4em\relax IEEE, 2017, pp. 74--79.

\bibitem{yao2019clustering}
Y.~Yao, X.~Zhao, Y.~Wu, Y.~Zhang, and J.~Rong, ``Clustering driver behavior
  using dynamic time warping and hidden markov model,'' \emph{Journal of
  Intelligent Transportation Systems}, pp. 1--14, 2019.

\bibitem{niu2019label}
X.~Niu, T.~Chen, C.~Q. Wu, J.~Niu, and Y.~Li, ``Label-based trajectory
  clustering in complex road networks,'' \emph{IEEE Transactions on Intelligent
  Transportation Systems}, 2019.

\bibitem{yeh2016matrix}
C.-C.~M. Yeh, Y.~Zhu, L.~Ulanova, N.~Begum, Y.~Ding, H.~A. Dau, D.~F. Silva,
  A.~Mueen, and E.~Keogh, ``Matrix profile i: all pairs similarity joins for
  time series: a unifying view that includes motifs, discords and shapelets,''
  in \emph{2016 IEEE 16th international conference on data mining
  (ICDM)}.\hskip 1em plus 0.5em minus 0.4em\relax IEEE, 2016, pp. 1317--1322.

\bibitem{zhu2016matrix}
Y.~Zhu, Z.~Zimmerman, N.~S. Senobari, C.-C.~M. Yeh, G.~Funning, A.~Mueen,
  P.~Brisk, and E.~Keogh, ``Matrix profile ii: Exploiting a novel algorithm and
  gpus to break the one hundred million barrier for time series motifs and
  joins,'' in \emph{2016 IEEE 16th international conference on data mining
  (ICDM)}.\hskip 1em plus 0.5em minus 0.4em\relax IEEE, 2016, pp. 739--748.

\bibitem{zhu2018matrix}
Y.~Zhu, C.-C.~M. Yeh, Z.~Zimmerman, K.~Kamgar, and E.~Keogh, ``Matrix profile
  xi: Scrimp++: time series motif discovery at interactive speeds,'' in
  \emph{2018 IEEE International Conference on Data Mining (ICDM)}.\hskip 1em
  plus 0.5em minus 0.4em\relax IEEE, 2018, pp. 837--846.

\bibitem{yeh2017matrix}
C.-C.~M. Yeh, N.~Kavantzas, and E.~Keogh, ``Matrix profile vi: Meaningful
  multidimensional motif discovery,'' in \emph{2017 IEEE International
  Conference on Data Mining (ICDM)}.\hskip 1em plus 0.5em minus 0.4em\relax
  IEEE, 2017, pp. 565--574.

\bibitem{bezzina2015safety}
D.~Bezzina and J.~Sayer, ``Safety pilot model deployment: Test conductor team
  report, usdot report no. dot hs 812 171,'' 2015.

\bibitem{wang2017much}
W.~Wang, C.~Liu, and D.~Zhao, ``How much data are enough? a statistical
  approach with case study on longitudinal driving behavior,'' \emph{IEEE
  Transactions on Intelligent Vehicles}, vol.~2, no.~2, pp. 85--98, 2017.

\bibitem{rakthanmanon2013addressing}
T.~Rakthanmanon, B.~Campana, A.~Mueen, G.~Batista, B.~Westover, Q.~Zhu,
  J.~Zakaria, and E.~Keogh, ``Addressing big data time series: Mining trillions
  of time series subsequences under dynamic time warping,'' \emph{ACM
  Transactions on Knowledge Discovery from Data (TKDD)}, vol.~7, no.~3, p.~10,
  2013.

\bibitem{bu2007wat}
Y.~Bu, T.-W. Leung, A.~W.-C. Fu, E.~Keogh, J.~Pei, and S.~Meshkin, ``Wat:
  Finding top-k discords in time series database,'' in \emph{Proceedings of the
  2007 SIAM International Conference on Data Mining}.\hskip 1em plus 0.5em
  minus 0.4em\relax SIAM, 2007, pp. 449--454.

\bibitem{grabocka2014learning}
J.~Grabocka, N.~Schilling, M.~Wistuba, and L.~Schmidt-Thieme, ``Learning
  time-series shapelets,'' in \emph{Proceedings of the 20th ACM SIGKDD
  international conference on Knowledge discovery and data mining}.\hskip 1em
  plus 0.5em minus 0.4em\relax ACM, 2014, pp. 392--401.

\end{thebibliography}
\end{document}